\PassOptionsToPackage{table}{xcolor}

\documentclass[]{jingdong}
\definecolor{jsonstr}{rgb}{0.5,0.5,0.5}

\usepackage{marvosym}
\usepackage{listings}
\usepackage{xcolor}
\usepackage{array}

\colorlet{jdhl}{jingdongred!8}

\lstdefinelanguage{json}{
    basicstyle=\ttfamily\tiny,
    showstringspaces=false,
    breaklines=true,                 %
    breakatwhitespace=false,
    frame=single,                    %
    rulecolor=\color{gray!40},
    backgroundcolor=\color{gray!5},  %
    literate=
     *{:}{{{\color{black}{:}}}}{1}
      {,}{{{\color{black}{,}}}}{1}
      {\{}{{{\color{black}{\{}}}}{1}
      {\}}{{{\color{black}{\}}}}}{1}
      {[}{{{\color{black}{[}}}}{1}
      {]}{{{\color{black}{]}}}}{1},
    morestring=[b]",
    stringstyle=\color{jsonstr},
}

\usepackage[utf8]{inputenc}
\usepackage[T1]{fontenc}

\usepackage{amsfonts}
\usepackage{amsmath}
\usepackage{amssymb}
\usepackage{booktabs}
\usepackage{tabularx}
\usepackage{xcolor}
\usepackage{enumitem}
\usepackage{inconsolata}
\usepackage{listings}
\usepackage{nicefrac}
\usepackage{url}
\usepackage{xspace}
\usepackage{wrapfig}

\crefformat{section}{\S#2#1#3}
\Crefformat{section}{\S#2#1#3}
\crefmultiformat{section}{\S#2#1#3}{ and \S#2#1#3}{, \S#2#1#3}{ and \S#2#1#3}
\Crefmultiformat{section}{\S#2#1#3}{ and \S#2#1#3}{, \S#2#1#3}{ and \S#2#1#3}
\crefrangeformat{section}{\S#3#1#4 to \S#5#2#6}
\Crefrangeformat{section}{\S#3#1#4 to \S#5#2#6}

\newcommand{\best}[1]{{\sffamily\bfseries #1}}
\newcommand{\second}[1]{\underline{#1}}

\definecolor{templatekeyword}{HTML}{E1251B}
\definecolor{templatestring}{HTML}{0B7A75}
\definecolor{templatecomment}{HTML}{6B7280}
\definecolor{templatecodebg}{HTML}{F7F7F8}

\lstdefinestyle{templatecode}{
  basicstyle=\ttfamily\small,
  columns=fullflexible,
  backgroundcolor=\color{templatecodebg},
  frame=single,
  rulecolor=\color{black!15},
  numberstyle=\tiny\color{gray},
  keywordstyle=\color{templatekeyword},
  commentstyle=\color{templatecomment},
  stringstyle=\color{templatestring},
  showstringspaces=false,
  tabsize=2,
  breaklines=true,
  breakatwhitespace=true,
  captionpos=b,
  xleftmargin=3.4pt,
  xrightmargin=3.4pt
}

\renewcommand{\authorfont}{\fontsize{9.5}{11.5}\selectfont}
\renewcommand{\affiliationfont}{\fontsize{9}{11}\selectfont}
\renewcommand{\contributionfont}{\fontsize{9}{11}\selectfont}

\title{JoyAI-VL-Interaction: \\ Real-Time Vision-Language Interaction Intelligence}

\author[*]{Dingyu Yao}
\author[*]{Junhao Zhou}
\author[*]{Chenxu Yang}
\author[*]{Chuanyu Qin}
\author[*]{Xiangyu Zeng}
\author[*]{Yifei Li}
\author[*]{Haowen Hou}
\author[]{Zheming Liang}
\author[]{Congcong Wang}
\author[]{Kaiwen Tuo}
\author[]{Jun Zhang}
\author[]{Yuhan Zhu}
\author[]{Yuhang Cao}
\author[]{Shenglong Ye}
\author[]{Shuai Xie}
\author[]{Shuhuan Gu}
\author[]{Haoyang Huang}
\author[*,\dagger]{Qingyi Si}
\author[\ddagger]{Nan Duan}
\author[\ddagger]{Jiaqi Wang}

\contribution[*]{Equal Contribution}
\contribution[\dagger]{Project Lead}
\contribution[\ddagger]{Corresponding Authors}
\affiliation[]{JD.com}

\checkdata[Repository]{\href{https://github.com/jd-opensource/JoyAI-VL-Interaction}{https://github.com/jd-opensource/JoyAI-VL-Interaction}}

\abstract{
Many moments in the real world do not wait for a user to ask. A fire starts on a security monitor, an expression flickers across a video call, or a product a viewer wants flashes by in a livestream. Once missed, the moment is gone, because the physical world does not pause. Yet today’s large models remain mostly turn-based by design: they answer only when addressed, and even video-call apps that appear interactive still operate as question-answer systems, reacting only when polled or prompted. We argue for a different paradigm: a model that is present in the world like a person. It continuously watches what is happening now, decides on its own whether to speak or stay silent, interacts in real time, and delegates to a background model when the problem is hard. This opens a "watch-and-do" mode of human-AI collaboration, expands AI assistance to far more real-life situations, and aligns naturally with the real-time demands of embodied intelligence. To advance interaction models and their adoption across domains, we make two fully open-sourced contributions. First, we release \textbf{JoyAI-VL-Interaction}, an 8B-scale, vision-first VL-interaction model that is proactive by design. The model makes the response decision internally, choosing each second to stay silent, respond, or delegate to a background model, and it excels at vision-triggered responsiveness and time awareness. We pair it with a transferable training recipe, from which capabilities we never trained for emerge, such as guiding a shopper through changing app screens or improvising a lecture from a slide deck. Second, we release a complete, deployable system built around that model. The system streams any ongoing video, from cameras, livestreams, or security feeds, into the model, making it genuinely present in the world. It supports hours of continuous video with sub-second latency. All other components are pluggable, including ASR/TTS modules, memory, visualization UI, and a background brain that can connect to any API, model, or agent.
Across six real-world streaming scenarios, human raters prefer JoyAI-VL-Interaction over the in-app video-call assistants of Doubao and Gemini by a wide margin on both quality and timing, winning 77.6\% against Doubao and 87.9\% against Gemini. It also attains the best average among comparable open models on 26 standard video benchmarks, leading every temporal-grounding and streaming benchmark without giving up general video understanding. To our knowledge, this is the first open, vision-driven interaction model released together with its training recipe, data, and complete deployable system.
With our repository, anyone can deploy an assistant that sees the present, speaks up on its own, and moves fluidly between the physical and digital worlds. 
Our goal is to move VLMs beyond turn-based dialogue toward real streaming interaction, openly and together.

}

\begin{document}
\maketitle

\section{Introduction}
\label{sec:introduction}
The moments that most need AI are often the very moments no one has time to ask for it. 
A toddler wanders toward a hot stove. The decisive play in a match is over before anyone has time to cheer. An aging parent living alone falls in the next room. What these moments share is not merely urgency, but timing: the instant that calls for a word, a cheer, or a steadying hand arrives before anyone thinks to ask, and disappears just as quickly.
Today’s large models are not built for such moments. Most remain fundamentally turn-based: they wait to be addressed, and respond only after they are asked.
This limitation is structural, not merely a matter of speed. A turn-based model cannot perceive ``now'' by construction: it opens its eyes only at the moment it is prompted. What these situations require instead is a model that continuously sees what is happening in the present and interacts with the world as events unfold.

Existing efforts do not truly solve this problem. One line of work is real-time omni models, such as GPT-Realtime-2 \cite{openai2026gptrealtime2} and Qwen3.5-Omni \cite{qwen2026omni}, which are genuinely end-to-end single models (audio and video in; speech out) with low latency and no stitched-together pipeline. 
However, their optimization target remains conversational turn-taking: responding as quickly as possible after the user has spoken. Fundamentally, they are still organized around dialogue. They wait for the user's turn, and are not designed to continuously watch a visual stream and decide on their own when the moment calls for speaking.
Another line of work is consumer video-call products. Doubao's in-app video call \cite{seed2_2026} chases proactivity by firing a background request every few seconds, but its core is still turn-based and its autonomy is locked to the polling cycle: an event is not observed until the next poll, so the system can never react sooner than that interval.
Gemini's in-app video call \cite{gemini31flashlive} goes even less far: it does not perform the background polling that Doubao relies on to monitor for events, and instead remains strictly in a one-question, one-answer interaction pattern.
A third line is research on streaming-video understanding \cite{qian2025dispider,qian2025streaming,videollm-online,wang2024mmduet,yan2026proactvl,guan2026videostreamingthinking,awesomevlmstreaming}, which mostly remains at the laboratory stage.  Its evaluation is also offline: even benchmarks built for streaming \cite{li2025ovobenchfarvideollmsrealworld,lin2024streamingbenchassessinggapmllms} score a reply after the fact, so they measure how well a model understands a stream but not whether it speaks at the right moment, or at all, when no one has asked. It rarely brings together the properties a real-world deployment demands~\cite{yao2026harnessingstreamingvideowild}, namely proactive response\cite{xia2025streamingvideoinstructiontuning,qian2025dispider,yan2026proactvl}, long-horizon memory \cite{yao2025timechatonline,zhang2026hermeskvcachehierarchical,zeng2026streamforest}, and real-time operation \cite{xu2025streamingvlmrealtimeunderstandinginfinite,wang2026acceleratingstreamingvideolarge}. 
It typically advances only one dimension at a time. In short, none of these systems meets the needs of real deployment: each is either turn-driven rather than event-driven, not genuinely present in the live world, or alive only inside a benchmark.

\textbf{JoyAI-VL-Interaction} is built for exactly this. We adopt the term \textit{interaction model} \cite{thinkingmachines2026interactionmodels} from Thinking Machines Lab (TML) and make its criterion explicit: an interaction model judges for itself, moment to moment, when the time is right to respond to the ongoing stream, the way a person does, choosing when to speak and when to stay silent rather than answering only when addressed. A turn-based model, however low its latency or native its architecture, cannot choose its own moment: it speaks only when a user's turn arrives. The difference is not how fast a model answers, but whether it decides for itself when answering is worth it. We and TML recognized at nearly the same time that interactivity is worth scaling as a capability of its own; where we part is the route to it. TML fuses speech and vision into one large model and shared a research preview; we make vision the first-class driver with speech as pluggable I/O, keep the model compact enough to run on local, low-cost hardware, and open the whole stack so anyone can deploy, reproduce, and build on it.

\textbf{Contribution 1 --- the VL-interaction model.}
Our central idea is simple: make when to act a learned, per-second decision of the model itself, with staying silent treated as a first-class action alongside speaking and delegating. We build this vision-first interaction model on our base vision-language model, JoyAI-VL 1.0 (\S\ref{sec:arch}). Speaking and staying silent are the root of proactivity and time awareness, knowing when a moment is worth a word and when it is not; delegating is how the model reaches for more capability when a problem outgrows real-time inference. We pair every second of the visual stream with the action it calls for, yielding time-aligned data, which we construct at scale (\S\ref{sec:data}); from this the model learns its interactivity, as a behavior of its own rather than something bolted on by external turn-detectors or activity heuristics. We open-source the model weights, the training recipe, and the data behind it. On 26 standard video benchmarks the model attains the best average among open models of comparable scale, ranking first on every temporal-grounding and streaming benchmark while staying within a point of the strongest baseline on general video understanding, so the interaction ability does not come at the cost of general competence (\S\ref{sec:vlbench}).
\begin{figure}[htbp]
    \centering
    \includegraphics[width=1\textwidth]{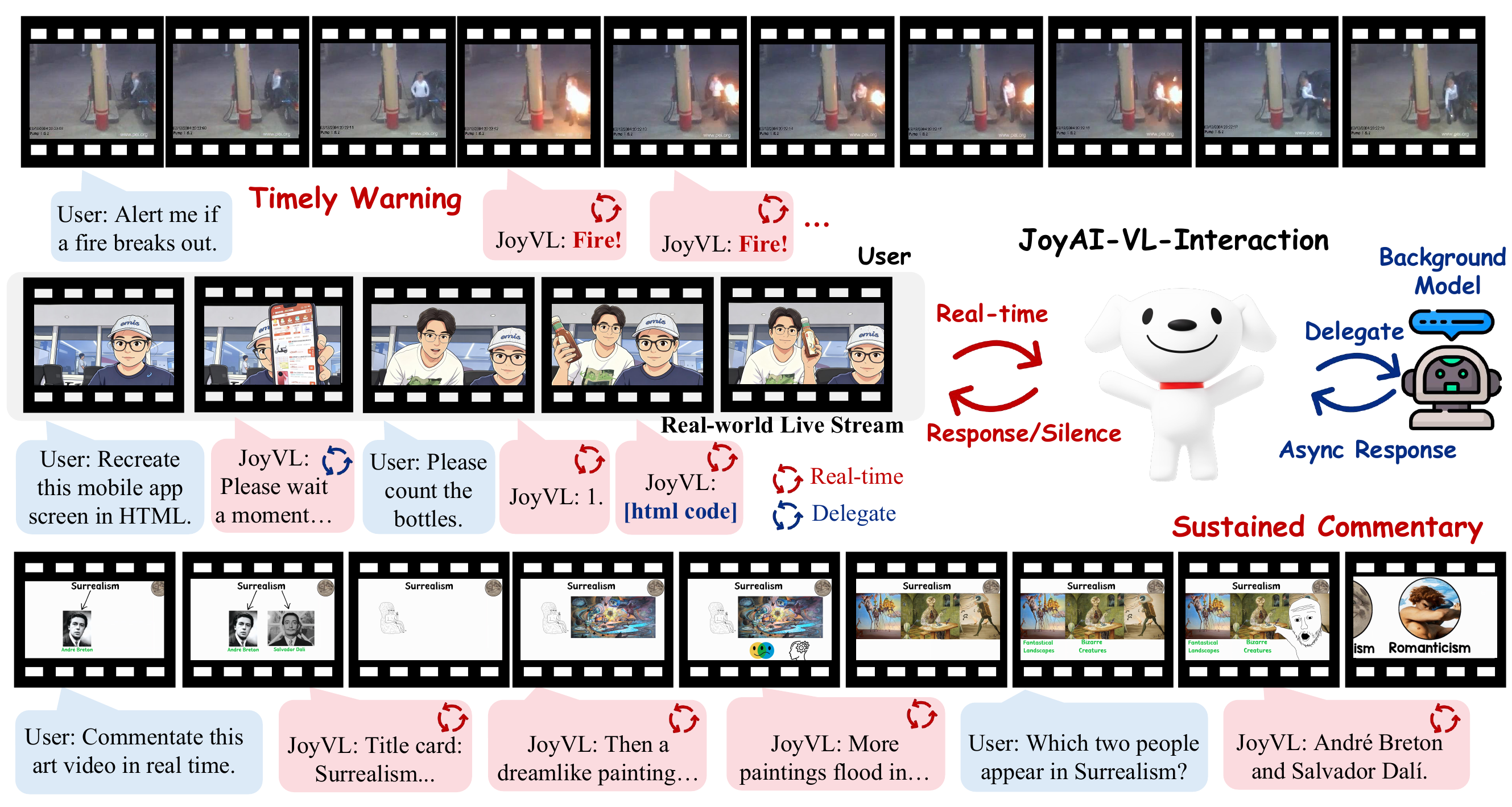}
    \caption{\textbf{JoyAI-VL-Interaction} processes a continuous video stream in real time, deciding at each step whether to respond, stay silent, or delegate complex tasks to a background model for asynchronous handling.}
    \label{fig:overview}
    \vspace{25pt}
\end{figure}

\textbf{Contribution 2 — the VL-interaction system.} We open-source a complete system that works out of the box: anyone can feed it a webcam or a livestream, and the interaction model immediately sees what is happening and interacts with the user in real time, genuinely present in the scene. Around the model, the system provides everything a deployment needs: ASR and TTS, a visualization UI, long-horizon memory, and a background bridge to any API, model, or agent the user brings, all served on standard vLLM to stay real-time over hours of continuous video. Among these, the interaction model alone decides whether and when to interact; the rest only transduce and orchestrate around it. Each ships with an open-source default, and the system provides the hooks to replace it, so a deployment can drop in its own ASR/TTS API or custom modules without rebuilding the stack. The system runs two concurrent loops, joined by the model's ``delegate'' action: a real-time loop with the user, and an asynchronous loop in which the model offloads a hard problem to a background brain while staying present with the user, folding the result back in when it returns. In a head-to-head human evaluation across six everyday scenarios, JoyAI-VL-Interaction is preferred over both Doubao's and Gemini's in-app video-call assistants by a wide margin on quality and timing, winning 77.6\% of comparisons against Doubao and 87.9\% against Gemini. Its strongest margins fall on the most time-critical settings: it wins every comparison in monitoring and alerting against both systems and never loses one in real-time translation or counting, exactly the event-driven, act-at-the-right-moment tasks that turn-based products structurally miss (\S\ref{sec:experiments}).

The value here is more than a better ``video assistant.'' When a model can be present and decide for itself what to watch and when to speak, human--AI collaboration shifts from ``submit a request and wait'' to ``watch-and-do,'' and AI can step into far more of everyday life, honoring embodied intelligence's premise that the physical world cannot be paused. The same capability is the substrate for a class of long-awaited products: a companion that is truly present, AI glasses that translate or caption what you see, an aide that serves as eyes for those who cannot, and uses no one has imagined yet. We release  JoyAI-VL-Interaction, the model, its training recipe and data, and the full system in the open, precisely so the community can build and explore all of this, and more, together. Our larger hope is to move the field from turn-based dialogue toward genuine streaming interaction, in the open.

\section{Related Work}

\subsection{Turn-based Models and Products}
\label{sec:product}
Almost all of today's models are \textit{turn-based}: they act only when a user takes a turn, answering once addressed and otherwise idle. This is a structural property, not a matter of speed. A turn-based system has no mechanism for reacting to an event in the world that arrives with no accompanying user utterance, however quickly it can reply once spoken to. Two families of recent systems push hard on responsiveness while keeping this turn-based core.

\paragraph{\textbf{Realtime and omni models.}} OpenAI's GPT-Realtime-2 reasons inside a single speech-to-speech loop \cite{openai2026gptrealtime2}, and Alibaba's Qwen3.5-Omni \cite{qwen2026omni} is a natively pretrained omni-modal model (text, audio, image, video) with real-time streaming and built-in turn-taking and interruption handling, the latter openly released. These remove the latency of cascaded stacks, but what they optimize is \textit{conversational turn-taking}: how quickly and naturally they answer once the user has spoken. Their interaction is organized around dialogue and waits for the user's turn; they are not built to watch a visual stream and decide, unprompted, that this is the moment to speak. Being open, real-time, and multimodal (as Qwen3.5-Omni already is) is therefore necessary but not sufficient for the setting we target.

\paragraph{\textbf{Consumer video-call products.}} In-app ``video call'' features are the most familiar real-time multimodal assistants and our most recognizable baseline, yet their cores remain turn-based. Doubao's runs on ByteDance's Volcano Engine conversational-AI stack, a configurable ASR/VLM/TTS pipeline with stop/turn detection, and its documentation makes the mechanism plain: while no one is speaking, the server only caches incoming video frames and does not send them to the model, so the application must periodically fire an \texttt{ExternalTextToLLM} trigger to obtain any analysis \cite{volcengine2026videounderstanding}; this characterization is based on the publicly available version of the stack, which may differ from the deployed product. ``Monitoring'' is thus an external clock bolted onto a turn-based model: reaction to an on-screen event waits for the next trigger and can never beat the polling interval. Gemini's video call goes less far still; in our use it answers only about the frame visible the instant a question is asked. In neither case does the decision of when to respond live in the model.

\subsection{Interaction Models}
A distinct, newer line breaks with the turn-based assumption by moving interactivity into the model.
\paragraph{\textbf{TML interaction models.}} Thinking Machines Lab recently named this an \textit{interaction model} and argued, as we independently and concurrently concluded, that scalable interactivity should come from the model itself rather than an external harness of turn detectors and activity heuristics. Their model handles audio, video, and text in real time and can speak unprompted, acting on the ongoing stream rather than waiting for a user's turn, and, as in our system, it is paired with an asynchronous background model for slower, harder reasoning, the two sharing context; it was released as a research preview \cite{thinkingmachines2026interactionmodels}. That two groups arrived at this approach at nearly the same moment we read as a sign that the move from turn-based to interactive models is a direction whose time has come. Within this shared interaction--background design, our aesthetic differs in two deliberate ways. On modality, rather than fusing audio and vision into the model alike, we keep speech (ASR/TTS) a pluggable module and make vision the model's intrinsic, primary modality for proactive interaction, fitting a watch-and-interact setting in which voice is interchangeable I/O. On scale, TML-Interaction-Small is a 276B-parameter mixture-of-experts model (12B active) that its authors describe as the smallest they can serve at the required latency, whereas we deliberately choose a compact $\sim$8B model (balancing interaction-grade responsiveness, local and low-cost deployability, and capability) so that it can be run, fine-tuned, and reproduced widely rather than only on large infrastructure.
\paragraph{\textbf{Full-duplex speech.}} For speech, Kyutai's MoshiRAG \cite{chien2026moshiragasynchronousknowledgeretrieval} is the prominent open instance: a real-time, full-duplex speech--text model in which both sides can speak and listen at once rather than strictly alternating, and the closest open precedent for putting interaction in the model. Its modality and scope (spoken conversation) set it apart from the visual, event-driven setting we address.

\subsection{Streaming Video Understanding}
A research line lets models process video as it arrives rather than after the fact, spanning streaming video LLMs, proactive response \cite{xia2025streamingvideoinstructiontuning,qian2025dispider,yan2026proactvl}, real-time inference \cite{xu2025streamingvlmrealtimeunderstandinginfinite,wang2026acceleratingstreamingvideolarge}, and long-horizon video memory \cite{yao2025timechatonline,zhang2026hermeskvcachehierarchical,zeng2026streamforest}. It is closest to our model contribution, but three gaps separate it from a deployable interaction model. It typically advances one property at a time (responsiveness, \textit{or} proactivity, \textit{or} memory) rather than the three together. It is usually measured on offline benchmarks, at times even in the manner of offline video understanding, which never tests reaction to live events under real-time constraints. And it stops at the model, without the surrounding system (serving, memory, transduction, delegation) that hours of sustained real-time presence demand. 

\subsection{Our Position.} Read along two axes, \textit{how} a model interacts and \textit{what} is released, JoyAI-VL-Interaction occupies a cell no prior work does. Against turn-based systems, native or product, the decision of when to act lives inside our model and is taken every second, event-driven, so reaction is bounded by inference rather than by a user's turn or a trigger clock. Against TML's interaction model, we make vision the first-class driver, the watch-and-interact setting rather than audio--video conversation, and we open-source not a research preview but the model, its training recipe, and a complete, composable, deployable system. Against MoshiRAG, we are vision-driven rather than speech-centric, reacting to events that carry no conversational turn at all. And against streaming-video research, we bring responsiveness, proactivity, and memory together in one model and run it inside a system built for sustained presence. To our knowledge, JoyAI-VL-Interaction is the first open, vision-driven interaction model released together with a complete deployable system: the point where all of these properties meet, rather than any one alone.

\section{JoyAI-VL-Interaction Model} \label{sec:model}

Figure~\ref{fig:overview} summarizes the model. JoyAI-VL-Interaction watches a streaming visual input and, at every second, takes one of three actions: speak to the user, stay silent and keep watching, or delegate a slower, harder task to an asynchronous background model whose result is folded back into the stream when ready. The first two actions form a real-time loop with the user; the third, an asynchronous loop with the background.

Two design points frame the rest of this section. First, the model learns a background-agnostic delegation protocol, so the background can be swapped for any external model, agent, or API; we cover the model side here and the system side in \S\ref{sec:system}. Second, the model does not process or generate speech itself: turning voice into text and back is left to pluggable ASR/TTS in the system, keeping the autonomous, vision-driven core decoupled from interchangeable I/O.

The rest of the section builds the model from this behavior outward: the base model and architecture (\S\ref{sec:arch}), the data that teaches when to speak, stay silent, and delegate (\S\ref{sec:data}), and the training recipe (\S\ref{sec:recipe}), which covers continue training, the training objective, reinforcement learning, and infrastructure.

\subsection{JoyAI-VL 1.0 and Architecture}
\label{sec:arch}

JoyAI-VL-Interaction is built on JoyAI-VL-1.0, %
where the language model is initialized from Qwen3-8B \cite{yang2025qwen3technicalreport}, the visual encoder is the Qwen3-VL ViT \cite{bai2025qwen3}, and the projection layer between them is trained from scratch. JoyAI-VL 1.0 is obtained through three stages, representation alignment, vision-language pre-training and post-train with On-Policy Distillation \cite{lu2025onpolicydistillation} and RL \cite{liu2026gdpogrouprewarddecouplednormalization,shao2024deepseekmathpushinglimitsmathematical}. At this point the model is a conventional, turn-based VLM. Its per-second interaction behavior, deciding each second whether to speak, stay silent, or delegate, is acquired afterward through the interaction-training recipe in \S\ref{sec:recipe}. JoyAI-VL-1.0 can optionally be further trained into a codec paradigm following AdaCodec~\cite{hou2026adacodecpredictivevisualcode}, which spends full ViT tokens only on reference frames and encodes predictable frames in between as compact P-tokens, reducing the per-frame token cost on long-horizon streams. We treat this as an optional extension and leave its evaluation to
future work: all experiments and demos in this paper use the non-codec version,  since the memory module and serving design described in \S\ref{sec:system} cover the full range of interaction scenarios we target, from live monitoring to multi-hour sessions.

\subsection{Data Construction for VL-Interaction }
\label{sec:data}

The applications we build for share a single demand: the assistant must be present and act on its own, at the moment that matters, with no one there to prompt it. A security feed where a flame has just appeared, a livestream where the item a viewer wants is flashing past, AI glasses that should caption or translate what is in front of you the instant it appears, a companion that should speak up only when there is something worth saying. In each case the right moment to act is set by the world, not by the user, and a model that waits to be addressed has already missed it. The data in this section is built to give the model exactly the abilities these settings demand: to watch a visual stream continuously and decide for itself, second by second, whether to speak, stay silent, or hand a hard problem to a heavier background model; to act on what it sees rather than wait to be asked; and to keep track of time, so that it knows not just what is happening but when. Teaching these abilities broadly, rather than for any single use case, is the goal of everything that follows. 

Concretely, at each one-second step the model takes one of three actions over the visual stream seen so far: it can stay silent and keep watching, emitting a \texttt{</silence>} token; speak, emitting a \texttt{</response>} token and a textual reply; or delegate a slower, harder subtask to an asynchronous background model. Delegation follows a fixed, background-agnostic protocol defined over text requests and results, so any external model, agent, or API can serve as the background (\S\ref{sec:system}). We teach these three decisions at one-second granularity, over and over, across as much of real life as we can capture.

\paragraph{\textbf{Capabilities the data covers.}}
To span this range, the data comprises more than 4M time-aligned streaming clips organized into six families. The first five are: (1) proactive alerting and anomaly detection on live feeds; (2) time-aligned question answering in backward, present, and forward timing; (3) counting and perception over time; (4) live commentary and narration; and (5) multi-turn casual chat in a variety of styles, ranging from everyday conversation over egocentric short video, to extended dialogue across long videos, to companionship-style exchanges. The sixth (6) cuts across all of the others: delegation episodes pair the ongoing scene with hard, off-stream subtasks—video-grounded knowledge questions, STEM problems, and video reasoning problems—that the model should route to the background rather than answer inline. A single per-second action format applies across every family, so each scenario jointly teaches the model when to speak, when to stay silent, and, where appropriate, when to delegate.

\paragraph{\textbf{Constructing time-aligned data.}}
Per-second supervision is what makes this behavior learnable, and it is costly to obtain at scale. Every example must be right along two axes: content, what the model says when it speaks, and timing, the exact second at which it should speak, stay silent, or delegate. Silence is a first-class label rather than the absence of one, since most steps in any stream should be silent and the model must learn to wait instead of speaking at every step. To get both axes right at scale, we run a multi-stage pipeline with dedicated verifier agents, and we deliberately favor quality over quantity. Each candidate example is checked at two levels: globally, over all input frames together with the full annotation, and locally, over the frames at the annotated timestamp and the reply tied to them. An example is admitted only if it passes both checks, and is discarded the moment it fails either. Each source is annotated along whichever axis is its bottleneck, and only examples that survive every check enter the corpus, so heterogeneous raw material is funneled into one clean, consistent per-second action format.

We tailor the construction to each family. For offline video-QA, the content of open-source data is already trustworthy, so the work is almost entirely about timing: a larger VLM marks when the evidence relevant to each question appears, and we form three timing types around it. In a backward example the question is asked after the evidence has appeared and the model responds at the moment the question arrives; in a present example the question is asked just as the evidence appears and the model responds at that moment; in a forward example the question is asked before the evidence appears and the model stays silent until the evidence arrives, then responds at that instant. For multi-turn casual chat, the casual nature of the exchange keeps the bar on content low, so we prioritize timing instead. Two VLM agents converse over a continuously playing video, one asking questions grounded in what it currently sees while the other watches the same stream and answers. To place each turn in time, we sample a time point at random and give the annotating model only the three frames around that point, which anchors every response to its timestamp and prevents large timing drift; a verifier agent then screens each exchange for grounding and quality. 

For commentary and narration, the commentator's authentic style is itself the value, so rather than synthesize it we collect open-source commentary and broadcast footage and recover the real spoken content with ASR, which preserves the natural cadence of when a person speaks and pauses and yields realistic silence labels. For counting, we run two passes over a counting-oriented source: the first keeps only videos in which objects reappear, and the second lays down the per-second labels. For perception over time, we insert time-conditioned content into ready-made multi-turn dialogues: for instance, requiring in the question that the answer be withheld for a precise interval before it is given, adding an instruction that increments a running count by one every $n$ seconds, or bounding the time window within which a question and its answer may occur; these rules yield exact, checkable timing without further manual annotation. 

 For alerting and anomaly detection, timing accuracy matters most, since an alert one second late describes a different moment, so we build this data from two sources. From open-source temporal-grounding annotations we convert directly into per-second labels, with a verification pass over the onsets. For web-collected videos, which carry no annotations, the pipeline proposes a candidate trigger window and tightens it through several stages: it filters videos to viable candidates, samples a window, generates and hard-filters a target by type, and applies a window-level verifier; a dense precheck then scans every 1 fps frame from the query time to the candidate trigger and discards the example if the target appeared earlier, so the alarm marks a genuine first onset, after which we localize the trigger to a single frame.

\paragraph{\textbf{Anatomy of a delegation episode.}}
Delegation is the most intricate behavior in the corpus, and the most distinctive: it is where the model learns to recognize the limits of real-time inference and hand off, mid-stream, to a heavier background brain without ever pausing the world it is watching. Each episode is built end to end, pairing a trigger (a subtask the model should not solve inline) with a written request and a returned result, and we synthesize these episodes three ways: (i) inserting ready-made pure-text hard problems such as STEM questions, where it is the difficulty rather than the modality that should send them to the background; (ii) converting open-source offline video-reasoning problems and using a large model to weave them into annotated multi-turn chat, so a genuinely hard visual question can surface naturally in the middle of a conversation; and (iii) synthesizing full episodes with a multi-role agent pipeline, in which a Planner scripts the episode, a Timestamp / Visual Verifier runs multi-level checks, over the whole clip and at the trigger frame, that the trigger and its timing are grounded in what is actually on screen, a Background agent produces the heavy answer, and a Foreground Rewriter turns it into a natural reply.

At the heart of the episode is the two-loop choreography it teaches. The instant the model decides to delegate, it does not go quiet. It first hands the user a brief holding reply (for example, ``let me look into that''), then emits a hidden delegate token and query that the user never sees, dispatching the hard subtask to the background while the visual stream keeps rolling. We then inject a random delay that simulates the background's variable reasoning time, and only after that delay is the result folded back in, at which point the model produces its formal answer in context. The delay is the whole point: by varying it, we force the model to stay present while a delegation is still pending, continuing to watch the scene, field new turns, and hold its silence when nothing is worth saying, rather than freezing until the background returns. This is the exact seam where the real-time loop with the user and the asynchronous loop with the background are stitched together, and the model learns to run both at once. Across all six families, however different their raw material, every construction is reduced to the same per-second labels of silence, response, and delegation and passes through the same multi-level verification, which is what lets such heterogeneous sources train a single, unified interaction policy.

\paragraph{\textbf{A transferable recipe, and signs of emergence.}}
Because the construction is defined by this common per-second format rather than by any one domain, the recipe transfers cleanly to new scenarios: adding a capability is as simple as supplying streams from that domain, annotated in the same per-second form and passed through the same verifiers, with no change to architecture or training objective. And the payoff scales. Even a modest amount of time-aligned data is enough for strong interaction behavior to emerge, a sign that the recipe is highly data-efficient and that time-aligned training is nowhere near saturated, with substantial headroom still ahead as the data grows. Most strikingly, capabilities we never explicitly trained for emerge on their own, including guiding a user through a purchase across changing app screens and improvising a full lecture from a slide deck. We read this as strong evidence that the model is acquiring a general watch-and-interact competence rather than memorizing scenario-specific tricks, and that the same recipe can grow the next generation of always-present assistants, from AI glasses that narrate what you see to companions and accessibility aides that serve as eyes for those who cannot. What we have seen so far is only the beginning. 

\subsection{Training Recipe of VL-Interaction}
\label{sec:recipe}

\paragraph{\textbf{Continue training.}}
Starting from JoyAI-VL 1.0, a single supervised training stage turns the base into an interaction model; reinforcement learning then follows below. We mix the time-aligned interaction data of \S\ref{sec:data} into a large pool of conventional turn-based data and fine-tune on the combined corpus. Keeping the conventional data in the mixture is what lets the model acquire per-second interaction without displacing the general video competence of the base, which the benchmark results of \S\ref{sec:vlbench} confirm. 
This echoes \S\ref{sec:data}: eliciting interaction is data-efficient, and time-aligned training is far from saturated.

\paragraph{\textbf{Training objective.}}
On time-aligned data, silence steps vastly outnumber the steps on which the model speaks, so the supervised targets are dominated by the \texttt{</silence>} token. Under a standard SFT loss this imbalance pushes the gradient toward continued silence and dilutes the signal for responding. Following our streaming-native training \cite{yao2026harnessingstreamingvideowild}, we therefore weight the assistant tokens by their role. Let $A$ be the supervised assistant-token positions and $C \subseteq A$ the control-token positions, where each control token is either \texttt{</silence>} or \texttt{</response>}. We assign $w^{\text{first}}_{\text{silence}}$ to the first \texttt{</silence>} in a run, $w^{\text{repeated}}_{\text{silence}}$ to a continued \texttt{</silence>}, and $w_{\text{response}}$ to \texttt{</response>}; every other position gets weight $1$. We set $w^{\text{first}}_{\text{silence}}=1$, $w^{\text{repeated}}_{\text{silence}}<1$ to down-weight silence continuations, and $w_{\text{response}}>1$ to up-weight response onsets. A delegation needs no separate weight, because it always rides inside a response. The objective is the normalized weighted cross-entropy
\begin{equation}
\mathcal{L}(\theta) = -\frac{1}{|A|} \sum_{j \in A} w_j \, \log p_\theta(y_j \mid y_{<j}).
\end{equation}
In practice we use $w^{\text{repeated}}_{\text{silence}}=0.4$ and $w_{\text{response}}=1.5$. We apply this weighted loss only to the time-aligned data; the conventional turn-based data is trained with the standard SFT loss.

\paragraph{\textbf{Reinforcement learning.}}
Continue training establishes the basic interaction behaviors, but supervised learning alone is insufficient for optimizing the finer dynamics of streaming interaction. In particular, a good interaction requires the model to make two related but distinct decisions: {when to act} and {what to say}. A response may arrive at the right moment but contain incorrect or ungrounded content, while a semantically correct answer may still be undesirable if it is produced too early, too late, or when the model should remain silent. We therefore adopt a decoupled reinforcement-learning design based on GRPO, where interaction control and response quality are evaluated with complementary reward signals. Timing-related supervision focuses on whether the model should respond, remain silent, or delegate, while content-related supervision evaluates the correctness, grounding, and usefulness of the generated response. This separation prevents one aspect of interaction quality from obscuring errors in another.

Streaming tasks also differ substantially in their temporal horizon. For relatively localized behaviors such as question answering, alerting, and delegation, we construct compact rollouts around informative moments in the stream, avoiding the large number of uninformative silent turns in a full video. For sustained behaviors such as narration and commentary, we further perform {windowed multi-turn self-rollout}. Within each sampled temporal window, the model interacts autoregressively with the stream, and its previous responses and silence decisions are retained in the context for subsequent turns. This exposes the policy to the states created by its own behavior and allows RL to optimize coherent interaction patterns, including response frequency, repetition, missed opportunities, and long periods of unnecessary silence, while keeping rollout cost manageable.

Finally, we combine {trajectory-level outcome supervision} with finer-grained {turn-level process feedback}. Stream-level rewards evaluate the overall quality of an interaction trajectory, including temporal coordination, grounding, coverage, non-repetition, and appropriate delegation. Turn-level judgments provide additional information about which individual responses contribute positively or negatively within an otherwise mixed-quality trajectory. The trajectory-level signal remains the primary group-relative optimization target, while local feedback is used to improve credit assignment within multi-response interactions. For behaviors whose quality is difficult to capture with deterministic rules, we use task-specific LLM judges. Together, these designs allow a unified RL stage to optimize both one-shot and sustained streaming interaction while preserving the distinct learning signals required by timing and content.

\paragraph{\textbf{Infrastructure.}}
We run our RL stage on EasyVideoR1 \cite{qin2026easyvideor1easierrlvideo}, an efficient reinforcement-learning framework for training vision-language models on video. Its reward system is task-aware, with unified routing and modular extension, so rewards for text, image, video, and our streaming interaction tasks are served from a single pipeline. 

\section{JoyAI-VL-Interaction System}\label{sec:system}
Around the interaction model of §\ref{sec:model} we build a complete, deployable system whose single principle is this: the model is the only component that decides when to speak and when to delegate, while everything else (ASR, TTS, memory, the background brain, the visualization UI) is transduction and orchestration placed around it. Every component ships with an out-of-the-box open-source default and can be swapped for whatever fits a deployment, including the model itself, which any VLM trained with our recipe (§\ref{sec:recipe}) can replace. This "decision in the model, the rest replaceable" split is the system's design thesis, and it buys two properties at once: the system is composable, so anyone can rebuild it for their own domain, and it is deployable today, because the periphery reuses standard infrastructure rather than a bespoke stack.

Compared with TML, which fuses speech into the model alongside vision, we keep the visual trigger, the model's own decision to act on what it sees, as the native, in-model capability, and treat speech as interchangeable I/O. Placing interchangeable I/O outside the model is therefore not a weakness but a deliberate decoupling of the autonomous core from the parts a deployment will want to choose for itself.

The rest of this section builds it up: the two concurrent loops it runs, the background bridge that joins them, and how that bridge closes a loop from seeing to acting (§\ref{sec:sys-loops}); the pluggable transduction, visualization, and user-supplied modules around the model (§\ref{sec:sys-plug}); the long-horizon memory that keeps it coherent over hours (§\ref{sec:sys-mem}); and the vLLM-native serving path and open release that sustain sub-second presence and put the system in anyone's hands (§\ref{sec:sys-serve}).

\begin{figure*}[ht]
  \centerline{\includegraphics[scale=0.5]{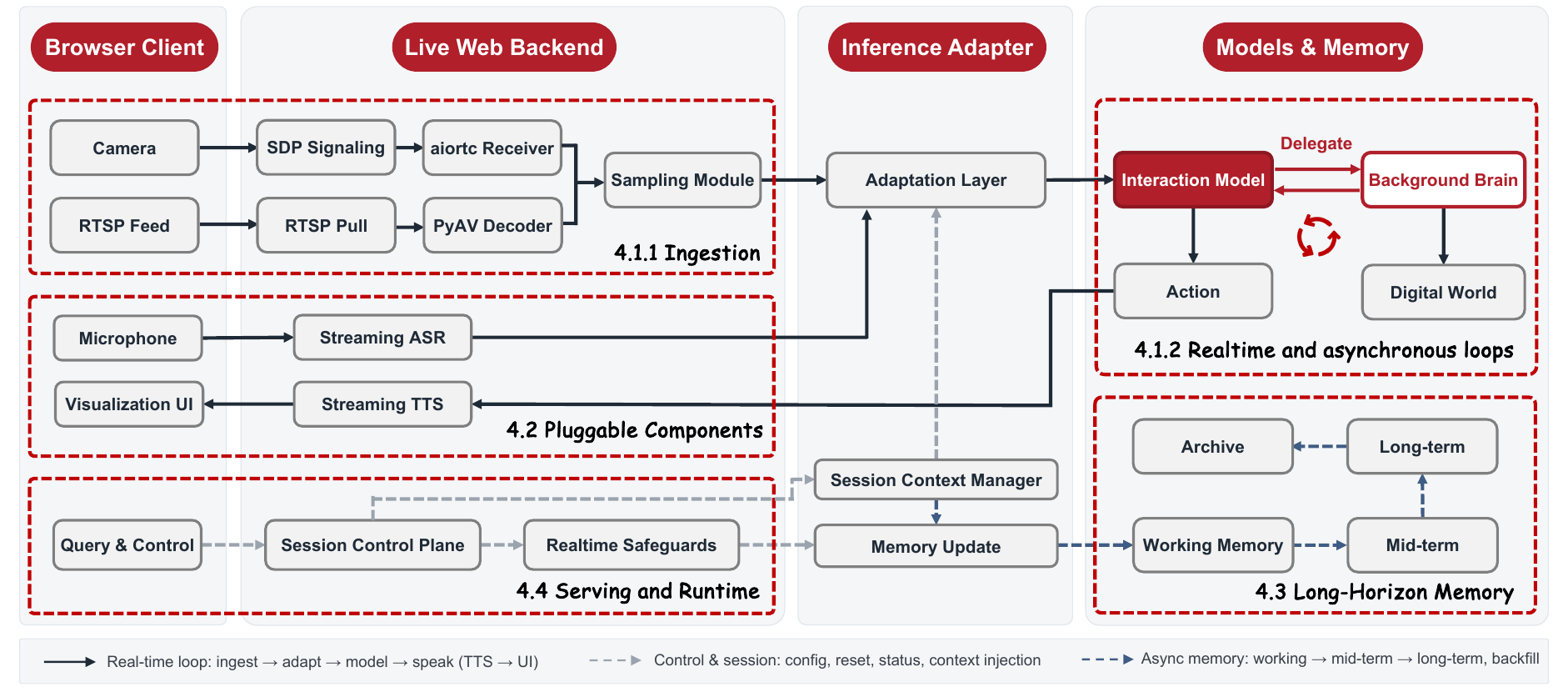}}
  \caption{Overview of the JoyAI-VL-Interaction System.}
  \label{fig:case}
\end{figure*}

\subsection{Two Concurrent Loops: from Seeing to Acting}\label{sec:sys-loops}
The interaction model runs two conversations at once: a real-time loop with the user and an asynchronous loop with the background brain, joined by the model's \texttt{delegate} action.

\paragraph{\textbf{Ingestion.}} The real-time loop begins at ingestion. A browser client pushes its camera stream to the server through WebRTC SDP negotiation, or submits an RTSP address that the server pulls; the two media paths are received and decoded server-side by aiortc and PyAV respectively, back into raw frames. A sampling module downsamples the stream at a fixed interval, 1\,Hz by default and configurable per scenario, to balance temporal detail against real-time latency. The sampled frames are converted to RGB, JPEG-encoded, and wrapped as Base64 data-URLs, then passed with the active user query into an adaptation layer that assembles an OpenAI-compatible multimodal Chat Completions request, the same request form our vLLM serving path expects (§\ref{sec:sys-serve}).

\paragraph{\textbf{The real-time and asynchronous loops.}} These frames are fed to the model every second, so it always acts on what is happening now rather than on a delayed batch. Together with optional microphone audio transduced by streaming ASR into an active query, they drive the model: at every second it takes one action, to speak, stay silent, or delegate; when it speaks, streaming TTS renders the reply and the visualization UI reflects its state. When it delegates, control passes to the asynchronous loop, in which a background model runs against the background brain to handle slower, more demanding tasks.

\paragraph{\textbf{The background bridge.}} The edge that \texttt{delegate} points to is a background bridge, and the brain behind it defaults to the user's own large-model API while remaining free to be any agent (such as Hermes Agent \cite{hermesagent} and OpenClaw \cite{openclaw} ) the user brings. The scaffold supports this natively and defines a fixed, background-agnostic protocol over text requests and results (§\ref{sec:data}). The bridge exposes a background-agnostic text contract: the foreground emits a tagged query, which the bridge normalizes into a task-id, delegated question, foreground note, and bounded frame snapshot. It runs an isolated model or agent asynchronously under a fixed timeout; expiry is reported as an error event and in-flight work is cancelled on session cleanup. Results return as started/ready/error events, with full artifacts kept off-context and only a bounded digest woven back into the interaction model. Because the background can be an agent, the bridge also closes a loop from seeing to acting: the interaction model observes the physical world and, when warranted, delegates a task the background agent carries out in the digital world. This is the watch-and-do premise of \S\ref{sec:introduction} made operational: see, decide, act, in one system. 

\subsection{Pluggable Components: Transduction, Visualization, and Your Own Modules}\label{sec:sys-plug}
\paragraph{\textbf{ASR and TTS.}} For speech, we ship a ready-to-run server built on open source ASR \cite{gao2023funasr} and TTS models \cite{du2025cosyvoice}, and users can swap in their own model or API to match their language and preferred voice. These modules only convert modalities; they take no part in the interaction decision. Because speech has duration while generation does not, we predefine a simple rule to keep the two in step: while the previous utterance is still being spoken, the model's next sentence is surfaced as text only and is not synthesized, so audio never piles up behind generation. For high-frequency settings such as live commentary, step-by-step guidance, and real-time subtitle translation, where output is near-continuous, we recommend running with TTS off, and we leave a fuller treatment of this regime to future work.

\paragraph{\textbf{Visualization UI.}} A built-in web interface, adapted from NVIDIA's open-source \texttt{live-vlm-webui}\footnote{\url{https://github.com/nvidia-ai-iot/live-vlm-webui}}, makes the running system visible and controllable. A left-hand configuration panel selects the video source—either a webcam or an RTSP livestream—and sets the sampling cadence, namely the number of frames per inference and the interval between inferences; it also offers a response mode that shifts the model's style from a default balance to more talkative or more aloof. The center pane shows the live stream the model is watching. The right-hand pane is the conversation: it holds the running exchange with the model, including earlier turns, reports per-response and average latency in real time, and accepts either typed messages or spoken input through the microphone (ASR); when the model speaks, its reply is played back automatically through TTS. The interface doubles as a debugging and trust surface for sensitive deployments, and it can be replaced with a custom front end.

\paragraph{\textbf{Bring your own modules.}} Beyond these defaults, the system is open to user-supplied modules that enrich what the assistant perceives or remembers. As one example, a face-recognition module can let the assistant remember who you are and recognize when a particular friend or family member walks into the scene, opening up more personal and engaging interactions. There is a larger reason we open all of this. Turn-based interaction is a long-settled range, refined over years, with a vast ecosystem of systems and optimizations built up around it; VL interaction models, by contrast, have barely begun. By releasing the full stack, we hope to open a first breach for interaction models among those established peaks, and we invite the community to explore with us what this still-young way of working with AI can become.

\subsection{Long-Horizon Memory}\label{sec:sys-mem}
Long-horizon presence needs a memory whose footprint does not grow without bound as a stream runs for hours, and whose contents form a shared context read by both the streaming model and the background brain. We meet this with a pluggable hierarchical memory built on our prior work~\cite{yao2026harnessingstreamingvideowild}, organized into tiers at increasing compression.
We organize the context into a three-tier hierarchy that partitions the stream into: (i) a \textit{short-term memory} retaining the most recent $T_s$ seconds as raw vision tokens; (ii) a \textit{mid-term memory} holding up to $M$ textual summaries of past short-term chunks, covering $T_m = M T_s$ seconds at moderate compression; and (iii) a \textit{long-term memory} storing up to $L$ aggressively compressed blocks, each consolidated from $M$ consecutive mid-term summaries and thus spanning $T_l = L M T_s$ seconds.
Alongside these visual tiers, a dialogue memory keeps past queries and answers coherent across time. The consolidation steps run asynchronously, ahead of each boundary, so they hide behind mainline inference and never stall the real-time loop, and together the tiers reach up to roughly two hours of context. Because each tier above the buffer is stored as text, the memory also forms a stable per-chunk prefix that the serving path of §\ref{sec:sys-serve} can cache and reuse; it is designed for exactly this, and it remains fully pluggable, swappable for any external memory a deployment prefers.

\subsection{Serving and Runtime}\label{sec:sys-serve}
\paragraph{\textbf{vLLM-native serving.}}
Hours of continuous video defeat the obvious serving strategies. Recomputing attention over the full history saturates the context window within a few hundred seconds; a sliding window bounds the context but breaks prefix reuse, since successive steps no longer share a common prefix, so an engine's prefix cache buys nothing and latency plateaus above the real-time budget; token pruning has the same problem. Most streaming-video methods inherit one of these and, built on plain Transformers, stay incompatible with efficient engines such as vLLM \cite{kwon2023efficientmemorymanagementlarge} and SGLang \cite{zheng2024sglangefficientexecutionstructured}, which is much of why they remain in the lab. We instead designed the memory system around prefix reuse so that vLLM serves it directly: the text memory of §\ref{sec:sys-mem} is prefilled once per chunk into a KV cache, and at every subsequent step only the newly observed frames and the previous reply are computed, while the memory and earlier in-chunk turns are reused without recomputation. Building the memory this way is precisely what lowers the bar to deployment: the system sustains over two hours of continuous video at sub-second end-to-end latency on standard vLLM. 

\paragraph{\textbf{Stateful sessions.}} The runtime is organized around server-side sessions. Control messages, configuration updates, session resets, and state synchronization, travel as JSON over WebSocket or HTTP and carry a \texttt{session\_id}, while the server keeps each session's video context, question-and-answer trajectory, and memory state. When a request is assembled, this maintained context is injected automatically into the Chat Completions request, so a single round of client-server communication carries only the instantaneous observation, and all long-horizon context management lives at the backend. This is also what makes the prefix reuse above possible: the injected context is precisely the stable per-chunk prefix the engine caches and reuses.

\paragraph{\textbf{Real-time robustness.}} To hold latency steady under load, the runtime places timeouts, reconnection, and cancellation across connection setup, the WebSocket channel, RTSP signaling, and the auxiliary audio and video channels, and a session-level lock keeps requests from piling up; when it must, it drops stale frames or backfills late results into the correct interaction history by inference sequence number. The adaptation layer routes each request by its model identifier to the corresponding VLM service, runs the mid-term and long-term memory consolidation of §\ref{sec:sys-mem} in parallel, and returns a standard JSON response.

\paragraph{\textbf{Everything is open.}} The model recipe, the background bridge, ASR/TTS, memory, orchestration, the visualization UI, and the vLLM serving path can all be replaced or extended in the repository. Releasing the model recipe and the complete system together is the whole point: from a single repository, anyone can stand up their own present, self-directed real-world assistant. 

\section{Experiments}
\label{sec:experiments}

We evaluate at three levels. \S\ref{sec:benchmark} is a product comparison: an interaction model is judged by the experience it delivers, so we place JoyAI-VL-Interaction against the two most widely used in-app video-calling assistants and let human raters decide. \S\ref{sec:svi} moves to SVI-Bench, which scores a complete interaction trajectory along five user-observable dimensions and lets us compare against recently released real-time and interactive models dimension by dimension. \S\ref{sec:vlbench} reports the conventional offline video suite, which tells us whether the interaction recipe preserved general video understanding.
\subsection{Real-World Streaming Evaluation}\label{sec:benchmark}
 Our primary evaluation is deliberately not an offline video-understanding benchmark. We do report those, in \S\ref{sec:vlbench}, where they answer a narrower question; here we measure JoyAI-VL-Interaction against the real, deployed products people would actually reach for, in the live, event-driven setting this paper is about. Concretely, we compare it head-to-head with the in-app video-call assistants of Doubao and Gemini, today's most mature and recognizable real-time multimodal products (\S\ref{sec:product}). We choose six everyday scenarios that fall squarely in the advantage zone of the interaction-model paradigm: settings that demand being present, acting unprompted at the right moment, and tracking events over time, precisely what a turn-based product, waiting to be addressed, is structurally unequipped to deliver however fast it answers once asked.

\paragraph{\textbf{Scenarios.}} The six scenarios are: (1) \textit{monitoring and alerting}, flagging an event the instant it occurs; (2) \textit{real-time counting} of objects or events over time; (3) \textit{real-time translation} of on-screen content; (4) \textit{time awareness}, acting on its own sense of elapsed time, for instance speaking once every few seconds on request or timing how long an activity lasts; (5) \textit{live commentary and guidance}, narrating or walking the user through an unfolding scene at the right moments; and (6) \textit{long-horizon memory}, answering about something seen far earlier in the stream. Underneath the six lie the three capabilities that define an interaction model: real-time operation, proactive response driven by what it sees, and long-horizon memory. The six are thus not an arbitrary set but a sampling of these axes, and the same capabilities reach far beyond this benchmark, into AI glasses, assistance for the blind, security monitoring, home robots, home and elder care, vision-grounded chat and companionship apps, live sports commentary, and delegation to background agents, among others. All are settings where being present and acting at the right moment is the whole task, the dimension on which turn-based products structurally fall short, so the six probe the paradigm gap itself rather than any single product feature.
The six scenarios comprise 58 cases in total: 10 each for monitoring, counting, translation, and time awareness, and 9 each for commentary and memory. 
The cases are drawn mostly from public footage on the web, with a few recorded by us.

\paragraph{\textbf{Baselines.}} Our baselines are the video-call features of the Doubao and Gemini apps, the products we single out in \S\ref{sec:product} as turn-based at the core, each evaluated as it is actually deployed to end users. We use the app versions current in late May and early June 2026, and drive each product with the same input under matched conditions. We run JoyAI-VL-Interaction through our own system (\S\ref{sec:system}) and compare it against each baseline separately, pairwise. 
The JoyAI-VL-Interaction system uses a three-tier memory with $T_s=100$\,s, $M=5$, and $L=15$ (\S\ref{sec:sys-mem}).
At evaluation time, our JoyAI-VL-Interaction system simulates offline videos as a live streaming source over RTSP, served by MediaMTX.
Since Doubao and Gemini expose no system-level API, we drive them through their apps: we play the same video to each and pose identical questions at matched timestamps.

\paragraph{\textbf{Protocol and metric.}} For every case, human raters score each system on two equally important axes: \textit{quality}, whether the response is correct, relevant, and well-formed; and \textit{timing}, whether it arrives at the right moment, neither premature nor late, and whether the system stays silent when nothing is worth saying. Each axis is rated on a three-level scale, good, fair, or poor, and a system's score for the case is the equal-weight average of its two axis ratings. The timing axis is the crux of the event-driven setting and the one turn-based products structurally struggle with (\S\ref{sec:introduction}). To reduce bias, system identities are hidden from raters and the presentation order is randomized. Five raters carry out the ratings, all educated to at least the university level and working as researchers in LLMs, and inter-rater agreement is high. For each case we then compare the two systems' weighted-average scores, which gives a win, tie, or loss for JoyAI-VL-Interaction. We report the per-scenario and overall \textit{win rate}, the fraction of comparisons it wins, against Doubao and against Gemini respectively, alongside the tie and loss rates; a tie counts as neither a win nor a loss.

\begin{table}[h]
\centering
\renewcommand{\arraystretch}{1.3}
\begin{tabular}{lccc}
\toprule
\textbf{Scenario} & \textbf{JoyAI-VL-Interaction} & \textbf{Tie} & \textbf{Doubao} \\
\midrule
Monitoring and alerting      & \textbf{100.0\%} & 0.0\%  & 0.0\%  \\
Real-time counting                & \textbf{70.0\%}  & 30.0\% & 0.0\%  \\
Real-time translation        & \textbf{80.0\%}  & 20.0\% & 0.0\%  \\
Time awareness               & \textbf{80.0\%}  & 10.0\% & 10.0\% \\
Live commentary and guidance & \textbf{55.6\%}  & 22.2\% & 22.2\% \\
Long-horizon memory          & \textbf{77.8\%}  & 22.2\% & 0.0\%  \\
\midrule
\textbf{Overall}             & \textbf{77.6\%}  & \textbf{17.2\%} & \textbf{5.2\%} \\
\bottomrule
\end{tabular}
\caption{Head-to-head human evaluation, JoyAI-VL-Interaction versus Doubao's in-app video-call assistant.}
\end{table}

\begin{table}[h]
\centering
\renewcommand{\arraystretch}{1.3}
\begin{tabular}{lccc}
\toprule
\textbf{Scenario} & \textbf{JoyAI-VL-Interaction} & \textbf{Tie} & \textbf{Gemini} \\
\midrule
Monitoring and alerting      & \textbf{100.0\%} & 0.0\%  & 0.0\%  \\
Real-time counting                & \textbf{100.0\%} & 0.0\%  & 0.0\%  \\
Real-time translation        & \textbf{100.0\%} & 0.0\%  & 0.0\%  \\
Time awareness               & \textbf{50.0\%}  & 40.0\% & 10.0\% \\
Live commentary and guidance & \textbf{100.0\%} & 0.0\%  & 0.0\%  \\
Long-horizon memory          & \textbf{77.8\%}  & 22.2\% & 0.0\%  \\
\midrule
\textbf{Overall}             & \textbf{87.9\%}  & \textbf{10.3\%} & \textbf{1.7\%} \\
\bottomrule
\end{tabular}
\caption{Head-to-head human evaluation, JoyAI-VL-Interaction versus Gemini's in-app video-call assistant.}
\end{table}

\subsection{Results and Case Study}

\paragraph{\textbf{Experimental Results.}}
Across both comparisons JoyAI-VL-Interaction is the preferred system by a wide margin, and in every scenario it wins more comparisons than it loses. Against Doubao it is preferred in 77.6\% of cases, ties 17.2\%, and loses only 5.2\%; against Gemini the margin is wider still, at 87.9\% wins, 10.3\% ties, and just 1.7\% losses. The two baselines, in other words, almost never come out ahead overall.

Our model's clearest advantage falls exactly on the vision-driven, time-critical scenarios that define the paradigm. In monitoring and alerting, the purest test of catching an event the instant it occurs, it wins every single comparison against both baselines (100\%). Real-time translation and fast counting are similarly lopsided: it sweeps both at 100\% against Gemini, and wins 80\% and 70\% against Doubao with no losses at all. These are precisely the settings where being present and acting at the right moment is the whole task, and they are where our margins are largest, which is the central claim of the benchmark borne out in the numbers. Long-horizon memory is also strong, at 77.8\% against each. Part of this margin reflects a structural limit of the baselines rather than a single weak answer: in video-call mode Doubao automatically hangs up after about five minutes with no voice input, and Gemini at around two minutes fifteen seconds, so in around half of the memory cases the question we ask falls past these cutoffs and neither baseline is even present to respond, scoring nothing. However, on the shorter cases that stay within their session limits, our model still wins or ties. We credit it to our long-horizon memory design: a mid-term and a long-term step that continuously compress, merge, and organize what the model has seen, so that information from far earlier in the stream is still at hand when a question finally arrives.

Where the baselines do gain ground, they do so for different reasons, and never on timing. Doubao's main foothold is live commentary and guidance, where it wins 22.2\% of cases and ties another 22.2\% against our 55.6\%. Its edge here is one of quality rather than timing: a larger model scale gives it broader knowledge, richer style, and more varied phrasing, which lands it in the comfort zone of some narration tasks. Its timing, in fact, is a weakness. On commentary, however, its timing works against it: the responses are temporally erratic, arriving too frequently in some passages and too sparsely in others, because its periodic external trigger has no means of judging when a remark is genuinely warranted. This corroborates our central design principle: deciding when to respond must be a native capability of the model, learned and judged internally, rather than a behavior supplied by an external trigger. Doubao therefore prevails on these cases only when its quality advantage is large enough to offset its weaker timing under the equal-weight scoring. Our own losses, in turn, are confined to quality: the model occasionally hallucinates during commentary, a limitation we attribute primarily to its parameter scale and regard as a principal direction for future work.
Gemini's single foothold is time awareness, the closest-contested axis against it, where it ties 40\% of cases and wins 10\%, holding our win rate to 50\%, while on every other scenario it fails to win a single comparison. The reason is specific to this scenario: some time-awareness cases are user-triggered, with the question asked only after the relevant moment has passed, so the real-time demand is low. On these ask-after-the-fact questions, Gemini's strong underlying model answers well on quality alone.
In both cases the baselines close the gap only where timing pressure is lowest, winning on raw answer quality or on questions that no longer require a split-second response, and they fall furthest behind exactly where a reply must land at the right moment.

\paragraph{\textbf{Case Study.}}
To make the quantitative gap concrete, we walk through six representative cases; the full side-by-side video comparisons are available in our blog.\footnote{\url{https://joyai-vl-video-future-academy-jd.github.io/JoyAI-VL-Interaction}} Each contrasts JoyAI-VL-Interaction with Doubao's and Gemini's in-app video-call assistants on identical input.

The \textit{Street Interview Translation} case (Demo~1: Subtitle translation) examines whether a single instruction remains active as new visual text appears. The task is to translate the interview's on-screen Chinese subtitles into English. JoyAI-VL-Interaction produces successive translations as the subtitles change, including the exchange about the interviewee's hometown and preferred city. The displayed subtitles contain both Chinese characters and pinyin, while the responses provide English renderings of the evolving dialogue. This sequence illustrates continuity in subtitle translation following a single request.

\clearpage

\begin{figure*}[t]
    \vspace{-40pt}
    \centering
    \includegraphics[width=0.9\textwidth]{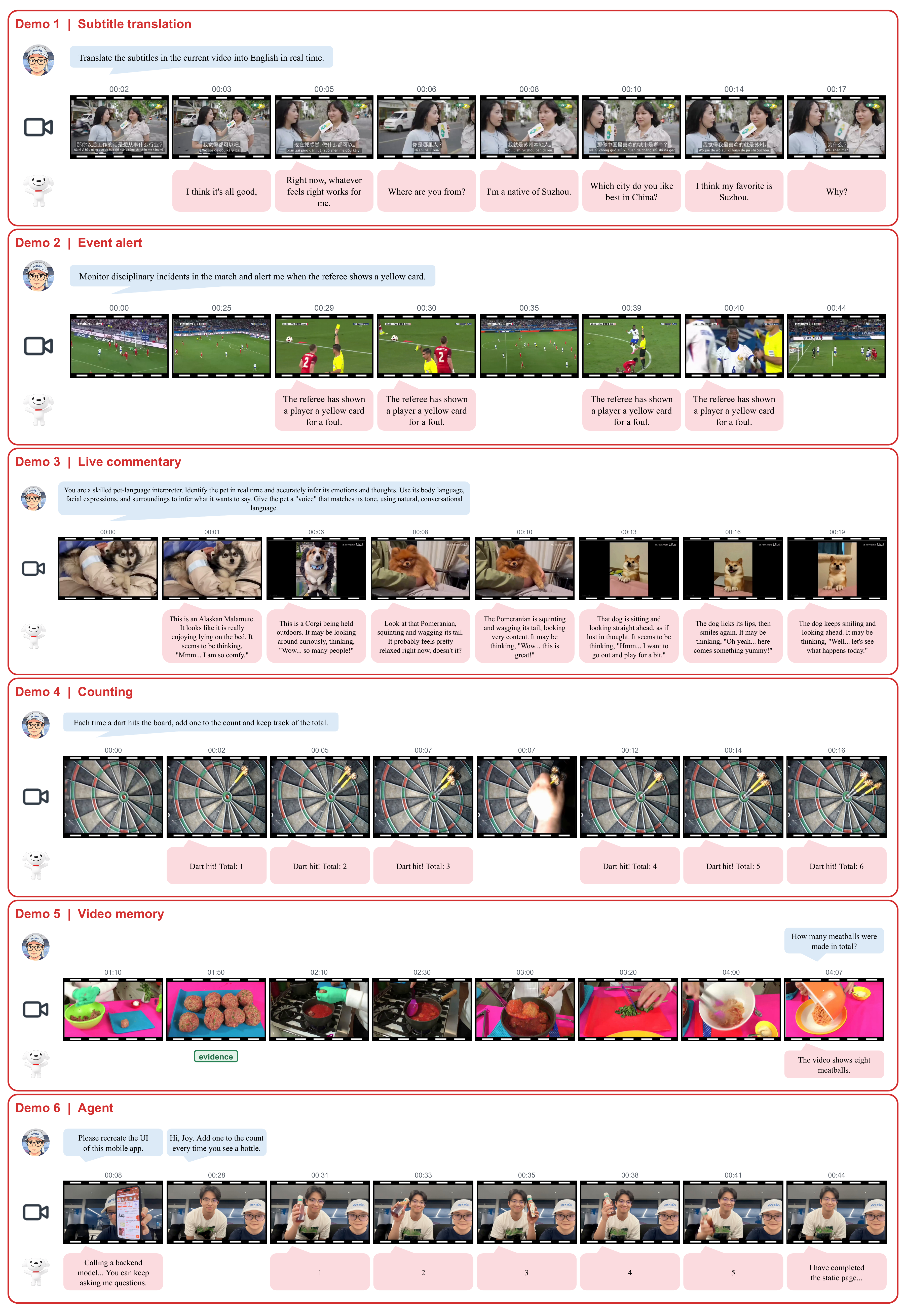} 
    \caption{Qualitative examples of JoyAI-VL-Interaction: subtitle translation, event alerts, live commentary, cumulative counting, video memory, and task delegation with concurrent bottle counting. Blue and pink bubbles denote user instructions and model responses, respectively.}
    \label{fig:demo_overall}
    \vspace{-10pt}
\end{figure*}

\clearpage

The \textit{Football Yellow-Card Alert} case (Demo~2: Event alert) examines visual monitoring under a standing instruction. The user asks the system to watch a football match and issue an alert when the referee shows a yellow card. JoyAI-VL-Interaction subsequently produces alerts associated with yellow-card situations in the displayed sequence. The case illustrates maintaining a user-specified monitoring condition as the scene evolves and generating relevant notifications without requiring the user to repeat the instruction.

The \textit{Pet Livestream Commentary} case (Demo~3: Live commentary) presents successive clips of different pets and asks the system to interpret their behavior in conversational language. JoyAI-VL-Interaction generates multiple comments as the animals and their actions change, describing posture and behavior while offering interpretations of apparent emotions and imagined thoughts. These interpretations are part of the requested commentary, rather than verified accounts of the animals' mental states. The sequence illustrates sustained commentary across changes in the visual scene.

The \textit{Real-time Dart Throw Counting} case (Demo~4: Counting) examines cumulative tracking of repeated visual events. After being instructed to increment the count whenever a dart hits the board, JoyAI-VL-Interaction produces successive totals from one to six in the displayed sequence. The running total continues across an intervening removal of darts from the board, even though the number of darts currently visible changes. This illustrates maintaining an accumulated event count across changes in the scene's visible state.

The \textit{Cooking Video Memory} case (Demo~5: Video memory) examines whether earlier visual information remains available for a later question. At the frame marked 04:07, the user asks how many meatballs were made in total. JoyAI-VL-Interaction answers that the video shows eight meatballs, consistent with the earlier frame marked 01:50 and highlighted as evidence. The relevant observation precedes the question by more than two minutes, with intervening cooking and serving steps. This example illustrates using earlier visual context to answer a subsequent question.

The \textit{Phone App Delegation with Concurrent Counting} case (Demo~6: Agent) illustrates continued live interaction during a delegated task. The user first asks the system to recreate a mobile application's visible interface. JoyAI-VL-Interaction announces a call to a backend model and indicates that the user can continue asking questions. The user then requests a count each time a bottle appears, and the system produces successive counts from one to five before reporting that the static page has been completed. The sequence illustrates coordination between a delegated interface-recreation task and ongoing visual interaction within the same session.

Two further settings move these capabilities off the desktop.\footnote{\url{https://github.com/jd-opensource/JoyAI-VL-Interaction/blob/main/demo}} In the \textit{Blind Assistance} setting, asked to watch a cup being filled and speak up before it overflows, the model stays silent through the pour and calls a halt on its own as the liquid nears the rim. Asked where the markers were left, it recovers both positions from earlier in the same stream, one in a compartment of the desk organizer and one lying flat on the desk. Asked to give unprompted warnings about the way ahead, it walks the user through an office area, flagging office chairs, table corners, a narrowing passage, and a tangle of cables and a power strip on the floor, each with a concrete instruction to slow down or pass on the right. A user who cannot see cannot ask about what they have not noticed, so none of these warnings could have been requested at the moment it was needed. The same capabilities appear in the \textit{AI Glasses} form, where a first-person feed is the model's natural input, across everyday scenes such as choosing clothes, crossing a street, locating a familiar product in a convenience store, and hearing a museum exhibit described.

\paragraph{\textbf{Emergent capabilities.}}
Two further cases are more striking still, in that they exercise capabilities the model was never trained for. In the \textit{Shopping App Guidance} case (03 App guidance in the blog), the user pursues a shopping goal while swiping through a phone's screens. JoyAI-VL-Interaction responds promptly and narrates in step with each swipe, guiding the user to the intended item, whereas neither Doubao nor Gemini reacts proactively or in real time, describing only a few of the screens first shown. Notably, our training data contains no app-interface video of any kind: the ability to follow a changing interface is one the model acquired on its own, generalizing to a domain it never saw.
In the \textit{Travel Scene Commentary} case (04 Live commentary in the blog), the system is asked to narrate once every four seconds. JoyAI-VL-Interaction holds to this cadence throughout and keeps the commentary substantively strong; Doubao narrates only the opening scene for some twenty seconds and ignores the four-second instruction entirely, while Gemini, though asked in Chinese, replies in English and comments only once. The task combines two abilities, timed action and live commentary, that never co-occur in our training data: the model composes them at inference time, an emergent crossing of capabilities rather than a pattern it was shown.

Across these cases the same factor separates JoyAI-VL-Interaction from systems many tens or even hundreds of times its size: timing. By making the decision of when to respond a native, learned capability of the model rather than a behavior imposed by an external trigger, JoyAI-VL-Interaction acts at the moment each event demands, where larger but turn-based or polling-based systems arrive late, stop after a single turn, or never engage. Moreover, its compact size imposes no ceiling on what it can ultimately handle: when a problem genuinely calls for a more powerful model, JoyAI-VL-Interaction does not attempt to answer it inline but delegates it, dispatching the hard subtask to a background model through an asynchronous request and folding the result back in once it returns, all while remaining present with the user in real time. A compact 8B model thus pairs real-time, well-timed presence with on-demand access to heavier reasoning, which we regard as the decisive reason it can outperform far larger models and mature products on the dimension that matters most in the streaming setting.

\subsection{Streaming Interaction Benchmark}\label{sec:svi}

SVI-Bench\footnote{\url{https://yidanhai.github.io/SVIbench-project-page/}} evaluates a complete interaction trajectory instead of individual answers, which lets us locate where an interaction breaks down. It scores five user-observable dimensions: proactive triggering (D1), whether the system speaks when an event calls for it; silence correctness (D2), whether it stays quiet when nothing does; latency (D3), whether the reply lands inside the required window; response correctness (D4); and delegation and long-horizon memory (D5). The first release holds 75 expert-authored items across nine interaction categories with 272 anchored dimension-level criteria, and every system is recorded through the same source video, query schedule and frontend while keeping its own streaming interface. Scoring is automatic but calibrated against expert labels: on the human pilot the judge matches 83.10\% of labels exactly and 97.18\% within one ordinal level, at a quadratic-weighted $\kappa$ of 0.78.

\begin{table}[!t]
\centering
\small
\setlength{\tabcolsep}{5pt}
\renewcommand{\arraystretch}{1.2}
\begin{tabular}{l c c c c c c}
\toprule
\textbf{System} & \shortstack{\textbf{D1}\\\textbf{Trigger}} & \shortstack{\textbf{D2}\\\textbf{Silence}} & \shortstack{\textbf{D3}\\\textbf{Latency}} & \shortstack{\textbf{D4}\\\textbf{Response}} & \shortstack{\textbf{D5}\\\textbf{Deleg./Mem.}} & \textbf{Overall} \\
\midrule
\rowcolor{jdhl}
JoyAI-VL-Interaction & \best{29.09} & 55.10 & \best{82.88} & \best{50.67} & \best{27.50} & \best{53.61} \\
Doubao Seed 2.1 Pro  & 17.27 & \best{91.84} &  2.74 & 40.00 &  7.50 & 21.22 \\
Mage-VL              & 15.45 & 81.63 & 14.38 & 25.33 &  7.50 & 19.50 \\
MOSS-VL-Realtime     & 22.73 & 59.18 & 23.97 & 32.67 & 15.00 & \second{25.44} \\
MiniCPM-o-4.5-9B     & 11.82 & 80.61 & 21.92 & 26.00 &  2.50 & 21.33 \\
\bottomrule
\end{tabular}
\caption{ Results on SVI-Bench, all values on a 0--100 scale. \best{Bold} = best, \second{underline} = second best on Overall. D1--D5 are raw per-dimension means over the items each dimension applies to; Overall couples D1 and D2 within each item by their minimum before averaging, so staying silent cannot by itself earn credit.}
\label{tab:svi}
\end{table}

Table~\ref{tab:svi} compares JoyAI-VL-Interaction against four deployed configurations, including the three most recent open models built for real-time or interactive use. JoyAI-VL-Interaction reaches an Overall of 53.61 while every other system stays below 26, and it leads four of the five dimensions. Silence correctness is the exception, where it places last at 55.10. The pairing of the two dimensions explains why. Doubao scores 91.84 on silence and 17.27 on triggering, and Mage-VL and MiniCPM-o-4.5 show the same shape, so their silence scores come from answering rarely rather than from judging when to answer. Our own gap between the two is 26 points, the smallest of the five. SVI-Bench anticipates this by coupling D1 and D2 through their per-item minimum before forming the Overall score, so silence alone earns no credit. JoyAI-VL-Interaction is therefore the only configuration that speaks when a response is due and stays quiet otherwise, while also ranking first on latency, response correctness, and delegation and memory. Removing latency from the aggregate leaves it first at 39.67.

\subsection{Video Benchmark Results}\label{sec:vlbench}

{We also evaluate JoyAI-VL-Interaction on standard video benchmarks. This evaluation serves two purposes. First, it tests whether converting a turn-based VLM into an interaction model preserves general video understanding. Second, it examines whether the interaction-oriented training brings additional benefits on tasks that depend more directly on temporal understanding and streaming video. The comparison set includes the open models positioned closest to our own, among them Mage-VL and MOSS-VL-Realtime, which also appear in \S\ref{sec:svi}. Of the four systems evaluated on both, JoyAI-VL-Interaction is the only one that leads on both axes, online interaction and offline understanding.
}

\begin{table}[!t]
\centering
\small
\setlength{\tabcolsep}{4pt}
\renewcommand{\arraystretch}{1.15}
\resizebox{\textwidth}{!}{%
\begin{tabular}{l l c c c c c c >{\columncolor{jdhl}}c}
\toprule
\rowcolor{white}
\textbf{Category} & \textbf{Benchmark} &
\shortstack{\textbf{MiniCPM-}\\\textbf{o-4.5}} &
\textbf{Mage-VL} &
\shortstack{\textbf{MOSS-VL-}\\\textbf{Realtime}} &
\shortstack{\textbf{MOSS-VL-}\\\textbf{Instruct}} &
\textbf{Qwen3-VL-8B} &
\textbf{InternVideo3} &
\shortstack{\textbf{JoyAI-VL-}\\\textbf{Interaction}} \\
\midrule
\multirow{6}{*}{\shortstack[l]{General\\Perception}}
 & TOMATO \cite{shangguan2025tomato}             & 30.66 & 36.05 & 38.68 & \second{40.63} & 35.70 & 38.21 & \best{41.85} \\
  & TVBench \cite{cores2024tvbench}               & 52.00 & 58.10 & 56.83 & 59.88 & 57.94 & \second{64.83} & \best{64.91} \\
 & MVBench \cite{li2024mvbench}                  & 64.31 & 66.73 & 61.18 & 65.62 & 68.70 & \best{74.65} & \second{73.70} \\
 & TempCompass \cite{liu2024tempcompass}         & 74.36 & 74.64 & 73.99 & \best{77.03} & \second{76.40} & 75.73 & 76.05 \\
 & PercTest \cite{patraucean2023perception}  & 67.52 & 67.48 & 65.88 & 68.51 & 72.70 & \best{78.83} & \second{75.89} \\
  & MotionBench \cite{hong2025motionbench}        & 59.56 & 60.08 & 60.25 & \best{62.94} & 58.40 & \second{61.75} & 60.05 \\
\midrule
\multirow{4}{*}{\shortstack[l]{Temporal\\Grounding}}
 & Charades-TL \cite{zhang2026timelens}         & 20.29 & 44.26 & 39.29 & 46.02 & 45.77 & \second{46.34} & \best{50.45} \\
 & ActivityNet-TL \cite{zhang2026timelens} & 13.44 & 42.99 & 36.60 & \second{46.06} & 45.59 & 43.80 & \best{52.12} \\
 & QVHighlight-TL \cite{zhang2026timelens}     &  9.02 & 54.63 & 50.52 & \second{59.53} & 59.08 & 59.02 & \best{65.03} \\
 & STVG \cite{li2025llava}                   &  0.25 &  8.74 & 10.87 & 13.77 & \second{15.00} & 13.84 & \best{17.56} \\
\midrule
\multirow{4}{*}{\shortstack[l]{Long\\Video}}
 & VideoMME V2 \cite{fu2026video}             & 27.31 & 23.50 & 26.63 & 27.97 & \best{32.50} & \second{31.94} & 31.84 \\
 & LongVideoBench \cite{wu2024longvideobench}    & 62.90 & 63.35 & 60.51 & 66.12 & \second{66.60} & 65.52 & \best{66.94} \\
 & VideoEval-Pro \cite{ma2025videoeval}       & 29.09 & 22.11 & 28.70 & 32.58 & 31.57 & \second{32.97} & \best{33.51} \\
 & EgoSchema \cite{mangalam2023egoschema}        & 59.40 & 68.00 & 57.40 & 62.80 & 69.80 & \best{77.00} & \second{75.60} \\
\midrule
\multirow{5}{*}{\shortstack[l]{Video\\Reasoning}}
 & Video-Holmes \cite{cheng2025video}      & \best{58.36} & 38.05 & 44.96 & 46.22 & 45.30 & \second{47.20} & 46.92 \\
 & VCRBench \cite{qi2025vcr}            & 38.20 & 35.40 & 35.98 & 39.65 & \second{42.70} & 41.20 & \best{43.04} \\
 & LongVideoReason \cite{chen2025longvila} & 79.67 & 73.09 & 79.44 & 79.08 & 78.40 & \second{81.08} & \best{82.26} \\
 & MMR-VBench \cite{zhu2026mmr}                 & 44.23 & 38.90 & 43.68 & 43.75 & 44.71 & \second{45.03} & \best{47.33} \\
  & VRBench \cite{yu2025vrbench}                  & \second{83.80} & 73.36 & 83.59 & 82.66 & 79.69 & \best{83.94} & 82.57 \\
\midrule
\multirow{4}{*}{\shortstack[l]{Video\\Knowledge}}
 & VideoMathQA \cite{rasheed2026videomathqa}     & 52.05 & 51.05 & 50.76 & 51.86 & 52.70 & \second{54.43} & \best{56.24} \\
 & VideoMMMU \cite{hu2026video}                  & 53.00 & 48.78 & 46.67 & 49.78 & \best{65.30} & 59.22 & \second{64.67} \\
 & MMVU$^{\dagger}$ \cite{zhao2025mmvu}                      & \best{60.20} & 53.10 & 50.20 & 52.40 & 58.70 & 57.90 & \second{60.00} \\
 & SciVideo \cite{deng2025scivideobench}         & \second{34.70} & 21.80 & 24.10 & 27.40 & 33.60 & \best{35.60} & \second{34.70} \\
\midrule
\multirow{3}{*}{\shortstack[l]{Streaming\\Video}}
 & OVOBench \cite{li2025ovobenchfarvideollmsrealworld} & \second{57.89} & 51.80 & 46.62$^{*}$ & 55.39 & 57.20 & 56.74 & \best{59.20} \\
 & OVBench \cite{huang2025online}               & 49.30 & 49.91 & 51.45$^{*}$ & 51.46 & \second{52.00} & 48.98 & \best{58.90} \\
 & ODVBench \cite{zeng2026streamforest}          & 47.68 & 53.84 & 54.35$^{*}$ & \second{59.45} & 56.10 & 53.17 & \best{68.60} \\
\midrule
\textbf{Average} & & 47.28 & 49.22 & 49.20 & 52.64 & 53.93 & \second{54.96} & \best{57.30} \\
\bottomrule
\end{tabular}}
\caption{ Video benchmark results across six capability blocks. Each entry uses the official metric of the benchmark, where higher is better. \best{Bold} = best, \second{underline} = second best. The \textbf{Average} row is the unweighted mean over all 26 benchmarks, and the block averages quoted in the text are unweighted means over the rows of each block. $^{*}$~MOSS-VL-Realtime is a streaming model, but is run here through its offline compatibility path, the only way to drive it under an offline protocol; these three scores are therefore diagnostic and are not comparable with the streaming results in its own report. $^{\dagger}$~MMVU uses the \texttt{mmvu-all} split.}
\label{tab:vlbench}
\end{table}

\paragraph{\textbf{Benchmarks.}}
{We evaluate on 26 public benchmarks, grouped into six capability blocks. General perception covers scene content and short-range motion. Temporal grounding requires locating an event within a video, in time and, for STVG, in space as well. Long video tests whether evidence distributed across a long input can be retrieved and combined. Video reasoning evaluates multi-step inference over video evidence rather than simple recognition. Video knowledge focuses on expert, scientific, and mathematical content that cannot be solved by perception alone.  Streaming video comes closest to our setting, as the model must answer about a video as it arrives, including information from moments that have already passed. Its scores are nonetheless computed after the fact, so we treat this block, and streaming benchmarks generally, as a measure of streaming comprehension; online interaction is what \S\ref{sec:benchmark} and \S\ref{sec:svi} measure.}

\paragraph{\textbf{Baselines.}} 
{We compare against six open models at a comparable scale. Three are general video-language models: Qwen3-VL-8B, InternVideo3-8B-Instruct, and MiniCPM-o-4.5. Qwen3-VL-8B is particularly relevant because JoyAI-VL 1.0 is initialized from the same backbone family, making it the closest control for studying the effect of our interaction-training recipe. The other three baselines focus more directly on streaming or real-time video understanding: Mage-VL, MOSS-VL-Realtime, and MOSS-VL-Instruct.
}

\paragraph{\textbf{Evaluation protocol.}} {We re-run all models under a unified evaluation harness rather than directly copying numbers from their reports. The benchmark definitions, prompts, answer parsers, and scorers are fixed across models so that the comparison reflects model differences rather than evaluation conventions. Each model follows its own official input and inference settings, since video sampling and preprocessing are part of a model's intended usage; Appendix~\ref{sec:reproduction} lists the per-model settings and compares our reproductions against the scores each release reports.}

\paragraph{\textbf{Results.}}
\noindent \textbf{Interaction training preserves general video understanding.} JoyAI-VL-Interaction achieves the best overall average among the seven models, with 57.30 compared with 54.96 for the strongest baseline. More importantly, its performance on conventional video-understanding tasks remains comparable to the strongest models in each block. On general perception, long video, video reasoning, and video knowledge, JoyAI-VL-Interaction obtains block averages of 65.41, 51.97, 60.42, and 53.90, respectively. These results are close to the strongest baseline in each category, with differences within about one point on three of the four blocks and a 1.33-point advantage on video knowledge.
This supports the first purpose of the evaluation: converting the base VLM into an interaction model preserves its general video competence.

\noindent \textbf{The main gains appear on temporal grounding and streaming video.} 
The performance gap becomes substantially larger when time itself is central to the task. JoyAI-VL-Interaction ranks first on all seven benchmarks in the temporal-grounding and streaming-video blocks. On average, it outperforms the next-best model by 4.93 points on temporal grounding and by 6.80 points on streaming video.
This pattern is also consistent relative to Qwen3-VL-8B, the closest control for our training recipe. The gains over Qwen3-VL-8B are largest on temporal grounding and streaming video, while improvements on the four conventional video-understanding blocks are much smaller.
The result is therefore not that interaction training makes the model uniformly stronger on every type of video task. Instead, it preserves general video understanding while producing its clearest gains on capabilities that depend on when an event happens and what information is available at a particular moment in a stream. This is consistent with the design of our time-aligned interaction data and the training objective in \S\ref{sec:recipe}, and mirrors the pattern observed in the real-world streaming evaluation of \S\ref{sec:benchmark}. We still count these gains as offline understanding. A strong OVOBench score shows that the model reads a stream well; whether it acts at the right moment within one is measured in \S\ref{sec:benchmark} and \S\ref{sec:svi}.

\subsection{Limitations}\label{sec:limitations}
A fair reading of our comparison has to start with scale. The two video-call assistants we evaluate against are powered by far larger models, Seed 2.0 behind Doubao and Gemini-3.1-flash-live behind Gemini, and both are mature products tuned over time against real users and use cases. JoyAI-VL-Interaction, by comparison, is a compact 8B-scale model. We therefore expect, and do not claim otherwise, that on general turn-based ability these products are stronger than ours: broader world knowledge, a more polished one-on-one chat experience, and greater robustness on complex or rarely seen inputs.

Our claim is narrower, and we think more telling. In the six event-driven scenarios this paper targets, the ones that turn on real-time presence, proactive interaction, and a sense of time, and that already carry rich deployment value, a far smaller model holds a natural advantage simply by being an interaction model rather than a turn-based one. That a compact, open model can outperform far larger and heavily optimized products precisely in this regime is, to us, the point: interactivity is worth scaling as a capability in its own right, and it begins to pay off well before enormous scale.

Both the data and the evaluation behind this release are, by deliberate choice, still at an early stage. On the data side, we have not yet tuned the mixture or scaled it to the volume we ultimately intend, and a further round of cleaning remains; one consequence is already visible in our results, where the comparatively sparse commentary data, together with the model's compact scale, is the main cause of the occasional hallucination during live narration (\S\ref{sec:experiments}).  The evaluation is similarly preliminary: the head-to-head study covers six scenarios and 58 human-rated cases against two products, and while \S\ref{sec:vlbench} adds 26 standard benchmarks, neither is the larger, more fine-grained study of live interaction we ultimately want, which no public benchmark yet measures. We release at this stage regardless, and the reason is the same for both: even with this early data and this limited evaluation, the model already exhibits interaction capabilities we never explicitly trained for, and watching them emerge convinced us, more than any single number could, that interactivity is a direction worth scaling in its own right. That conviction is exactly why we would rather put the model, the recipe, and the system into the community's hands now than hold them back for a polished data recipe and a larger benchmark: we are eager to scale this direction together rather than alone. A subsequent version will tune the data mixture, scale the corpus, and report a broader and more systematic evaluation; in the interim, it is the data-construction methodology and the approach itself, rather than this particular corpus or benchmark, that we expect others can most readily adapt and extend. These limitations also chart the road ahead: closing the gap in general ability and robustness, widening the set of scenarios, and pushing further on what an interaction model can do.

\section{Conclusion}\label{sec:conclusion}
We set out to move large models past the turn-based paradigm, in which a model opens its eyes only when it is addressed, toward genuine streaming interaction, in which the model is present in the world and decides for itself when to act. To that end we release, to our knowledge, the first open vision-driven interaction model, together with everything needed to build on it: the time-aligned data, the training recipe, the 8B model, and a complete, deployable system.

At the center is a single conviction, that interactivity should scale as a capability of the model itself rather than be simulated by a harness around it. Scaling interactivity means scaling three things at once: the model's freedom to speak on its own when a moment is worth a word, its ability to interact in real time, and its sense of elapsed time. As these grow, a model behaves less like a tool that waits to be queried and more like a participant that is simply present, watching what is happening and responding as a person would. We see this as a direction of scaling in its own right, alongside the scaling of intelligence, and one that brings models closer to being genuinely present in human settings.

Across the streaming scenarios we study, we already see early signs that an interaction model carries natural advantages for real deployment, from monitoring and live narration to step-by-step guidance and companionship. The moment we are reaching for is an everyday one: you come home worn out after a long day, and before you have said anything, a quiet voice notices and offers, ``I can see you're tired; today must have been hard on you.'' Presence like that, given unasked, is what an interaction model makes possible and a turn-based one, waiting to be addressed, never can. Much remains open in both the model and the system, and we have released the whole stack openly to lower the barrier and accelerate this shift. Turn-based models, products, and optimizations have been built up over many years, while vision-driven interaction models are only beginning; if this release opens even a first foothold for them among those long-settled peaks, and draws others into building it with us, it will have done its job. We invite the community to explore, with us, what a model that is truly present in the world can become. 
\section{Future Work}

The direction we are most eager to push forward is always-present assistance in the physical world, and two such settings have already moved past the idea stage in our own trials: AI glasses and assistance for the blind and visually impaired.\footnote{\url{https://github.com/jd-opensource/JoyAI-VL-Interaction/blob/main/demo}}
Our preliminary results in both are already promising, and we will keep optimizing along this direction to build a product dependable enough to be trusted every day: one that sees the world for those who cannot, turning AI from a service desk that answers only when asked into a companion that senses how you are and knows when a word is worth saying.

Reaching that point will take progress on the model and the paradigm as much as on the product, and several promising directions remain for extending JoyAI-VL-Interaction and the broader vision-driven interaction paradigm. One such direction is to explore real-time agentic tool-use capabilities, such as video replay, to enhance the perception of fast motion and fine-grained visual details while enabling more efficient and effective analysis of long streaming memories involving past video sequences. Another direction is to incorporate stronger spatial intelligence into streaming video understanding, extending the paradigm to egocentric wearable devices and embodied agents. Persistent spatial memory, 3D scene understanding, and spatially grounded reasoning could enable more effective perception and interaction in dynamic physical environments, broadening the scope of vision-driven interaction toward persistent and situated agents.

\bibliographystyle{assets/plainnat}
\bibliography{references}
\newpage
\section{Appendix}
\subsection{Training Data Example}
\begin{lstlisting}[language=json,caption={Example of monitoring and alerting.}, label={lst:example1}]
      {
        "role": "user",
        "content": "Alert me whenever a physical confrontation happens. \n<0.0 seconds>\n<image>\n<image>"
      },
      {
        "role": "assistant",
        "content": "</silence>"
      },
      {
        "role": "user",
        "content": "<1.0 seconds>\n<image>\n<image>"
      },
      {
        "role": "assistant",
        "content": "</silence>"
      },
      {
        "role": "user",
        "content": "<2.0 seconds>\n<image>\n<image>"
      },
      {
        "role": "assistant",
        "content": "</silence>"
      },
      {
        "role": "user",
        "content": "<3.0 seconds>\n<image>\n<image>"
      },
      {
        "role": "assistant",
        "content": "</silence>"
      },
      {
        "role": "user",
        "content": "<4.0 seconds>\n<image>\n<image>"
      },
      {
        "role": "assistant",
        "content": "</response> A physical confrontation is happening."
      },
      {
        "role": "user",
        "content": "<5.0 seconds>\n<image>\n<image>"
      },
      {
        "role": "assistant",
        "content": "</response> A physical confrontation is happening."
      },
      {
        "role": "user",
        "content": "<6.0 seconds>\n<image>\n<image>"
      },
      {
        "role": "assistant",
        "content": "</response> A physical confrontation is happening."
      },
      {
        "role": "user",
        "content": "<7.0 seconds>\n<image>\n<image>"
      },
      {
        "role": "assistant",
        "content": "</response> A physical confrontation is happening."
      },
      {
        "role": "user",
        "content": "<8.0 seconds>\n<image>\n<image>"
      },
      {
        "role": "assistant",
        "content": "</response> A physical confrontation is happening."
      },
      {
        "role": "user",
        "content": "<9.0 seconds>\n<image>\n<image>"
      },
      {
        "role": "assistant",
        "content": "</response> A physical confrontation is happening."
      },
      {
        "role": "user",
        "content": "<10.0 seconds>\n<image>\n<image>"
      },
      {
        "role": "assistant",
        "content": "</silence>"
      },
      ...
\end{lstlisting}

\newpage
\begin{lstlisting}[language=json,caption={Example of multi-turn casual chat.}, label={lst:example2}]
      ...
      {
        "role": "user",
        "content": "Is the person cleaning the desk?\n<7.0 seconds>\n<image>\n<image>\n<image>\n<image>"
      },
      {
        "role": "assistant",
        "content": "</response> Yep, definitely cleaning the desk! Getting all the dust off that pink surface."
      },
      {
        "role": "user",
        "content": "<8.0 seconds>\n<image>\n<image>\n<image>\n<image>"
      },
      {
        "role": "assistant",
        "content": "</silence>"
      },
      {
        "role": "user",
        "content": "<9.0 seconds>\n<image>\n<image>\n<image>\n<image>"
      },
      {
        "role": "assistant",
        "content": "</silence>"
      },
      {
        "role": "user",
        "content": "<10.0 seconds>\n<image>\n<image>\n<image>\n<image>"
      },
      {
        "role": "assistant",
        "content": "</silence>"
      },
      {
        "role": "user",
        "content": "<11.0 seconds>\n<image>\n<image>\n<image>\n<image>"
      },
      {
        "role": "assistant",
        "content": "</silence>"
      },
      {
        "role": "user",
        "content": "<12.0 seconds>\n<image>\n<image>\n<image>\n<image>"
      },
      {
        "role": "assistant",
        "content": "</silence>"
      },
      {
        "role": "user",
        "content": "What are they doing with the pillow?\n<13.0 seconds>\n<image>\n<image>\n<image>\n<image>"
      },
      {
        "role": "assistant",
        "content": "</response> They're picking up the pillow and moving it around, maybe fluffing it or putting it somewhere."
      },
      {
        "role": "user",
        "content": "<14.0 seconds>\n<image>\n<image>\n<image>\n<image>"
      },
      {
        "role": "assistant",
        "content": "</silence>"
      },
      {
        "role": "user",
        "content": "<15.0 seconds>\n<image>\n<image>\n<image>\n<image>"
      },
      {
        "role": "assistant",
        "content": "</silence>"
      },
      {
        "role": "user",
        "content": "<16.0 seconds>\n<image>\n<image>\n<image>\n<image>"
      },
      {
        "role": "assistant",
        "content": "</silence>"
      },
      {
        "role": "user",
        "content": "<17.0 seconds>\n<image>\n<image>\n<image>\n<image>"
      },
      {
        "role": "assistant",
        "content": "</silence>"
      },
      {
        "role": "user",
        "content": "Why is the person holding keys now?\n<18.0 seconds>\n<image>\n<image>\n<image>\n<image>"
      },
      {
        "role": "assistant",
        "content": "</response> Hmm, they seem to be fiddling with the keys on the desk. Maybe checking or organizing them?"
      },
      ...
\end{lstlisting}

\newpage
\begin{lstlisting}[language=json,caption={Example of background model delegation.}, label={lst:example3}]
      ...
      {
        "role": "user",
        "content": "<29.0 seconds>\n<image>\n<image>\n<image>\n<image>"
      },
      {
        "role": "assistant",
        "content": "</silence>"
      },
      {
        "role": "user",
        "content": "What are those two guys doing near the boat?\n<30.0 seconds>\n<image>\n<image>\n<image>\n<image>"
      },
      {
        "role": "assistant",
        "content": "</response> Looks like they're walking alongside it, probably supervising or guiding as it gets lifted."
      },
      {
        "role": "user",
        "content": "<31.0 seconds>\n<image>\n<image>\n<image>\n<image>"
      },
      {
        "role": "assistant",
        "content": "</silence>"
      },
      {
        "role": "user",
        "content": "<32.0 seconds>\n<image>\n<image>\n<image>\n<image>"
      },
      {
        "role": "assistant",
        "content": "</silence>"
      },
      {
        "role": "user",
        "content": "<33.0 seconds>\n<image>\n<image>\n<image>\n<image>"
      },
      {
        "role": "assistant",
        "content": "</silence>"
      },
      {
        "role": "user",
        "content": "<34.0 seconds>\n<image>\n<image>\n<image>\n<image>"
      },
      {
        "role": "assistant",
        "content": "</silence>"
      },
      {
        "role": "user",
        "content": "<35.0 seconds>\n<image>\n<image>\n<image>\n<image>"
      },
      {
        "role": "assistant",
        "content": "</silence>"
      },
      {
        "role": "user",
        "content": "<36.0 seconds>\n<image>\n<image>\n<image>\n<image>"
      },
      {
        "role": "assistant",
        "content": "</silence>"
      },
      {
        "role": "user",
        "content": "<37.0 seconds>\n<image>\n<image>\n<image>\n<image>"
      },
      {
        "role": "assistant",
        "content": "</silence>"
      },
      {
        "role": "user",
        "content": "Why does this boat hoist use a four-point sling configuration with spreader bars rather than lifting from a single central point, and how does the load distribution prevent hull deformation in fiberglass vessels?\n<38.0 seconds>\n<image>\n<image>\n<image>\n<image>"
      },
      {
        "role": "assistant",
        "content": "</response> Let me check the engineering principles behind that hoist design. </delegation> Explain the structural engineering principles of four-point boat lifting slings, load distribution, and hull deformation prevention in fiberglass boat hoists."
      },
      {
        "role": "user",
        "content": "<39.0 seconds>\n<image>\n<image>\n<image>\n<image>"
      },
      {
        "role": "assistant",
        "content": "</silence>"
      ...
\end{lstlisting}

\subsection{Reproduction Alignment}\label{sec:reproduction}

Each model follows its own official inference setting, listed in
Table~\ref{tab:settings}. Everything else is shared: the benchmark definitions,
prompts, answer parsers and scorers are fixed across the table.

\begin{table}[h]
\centering
\small
\renewcommand{\arraystretch}{1.15}
\begin{tabularx}{\textwidth}{@{}l X@{}}
\toprule
\textbf{Model} & \textbf{Inference setting} \\
\midrule
Qwen3-VL-8B            & Official Qwen3-VL setting: 2\,FPS, up to 2048 frames, 224K visual tokens \\
JoyAI-VL-Interaction   & Official Qwen3-VL setting: 2\,FPS, up to 2048 frames, 224K visual tokens \\
MiniCPM-o-4.5          & Released inference code: 1\,FPS, up to 64 frames, \texttt{model.chat}, audio/TTS/thinking off \\
InternVideo3-8B-Instruct & OpenGVLab Transformers path: 4\,FPS, 256--2048 frames per benchmark, deterministic decoding \\
MOSS-VL-Instruct       & Native \texttt{offline\_batch\_generate}: 1\,FPS, up to 256 frames \\
MOSS-VL-Realtime       & Offline compatibility path (see Table~\ref{tab:vlbench}) \\
Mage-VL                & HEVC \texttt{tc32} codec path: 32 canvases, 150K pixels \\
\bottomrule
\end{tabularx}
\caption{Per-model inference settings used for Table~\ref{tab:vlbench}.}
\label{tab:settings}
\end{table}

All scores in Table~\ref{tab:vlbench} are our own runs. Where a model's
release reports scores on benchmarks we also run, Table~\ref{tab:repro}
compares the two. LVBench is the largest gap for both models and is not one of
the 26 benchmarks in Table~\ref{tab:vlbench}. Our scores run slightly below the
published ones throughout, the direction to expect when prompts and scorers are
held fixed across models rather than tuned per model. Two baselines cannot be
checked this way: MOSS-VL-Instruct's release does not disclose the
preprocessing runner behind its published numbers, and Mage-VL publishes its
model-side codec settings but reports through \texttt{lmms-eval} rather than
the prompts and scorers used here.

\begin{table}[h]
\centering
\small
\renewcommand{\arraystretch}{1.15}
\begin{tabular}{l l c c c}
\toprule
\textbf{Model} & \textbf{Benchmark} & \textbf{Ours} & \textbf{Published} & \textbf{$\Delta$} \\
\midrule
\multirow{5}{*}{InternVideo3-8B-Instruct}
 & Video-MME      & 73.48 & 73.80 & $-0.32$ \\
 & MVBench        & 74.65 & 75.00 & $-0.35$ \\
 & MLVU           & 77.87 & 77.30 & $+0.57$ \\
 & LongVideoBench & 65.52 & 66.80 & $-1.28$ \\
 & LVBench        & 52.41 & 55.70 & $-3.29$ \\
\cmidrule(lr){1-5}
\multicolumn{4}{l}{\textit{mean absolute difference}} & \textit{1.16} \\
\midrule
\multirow{5}{*}{MiniCPM-o-4.5}
 & Video-MME      & 69.19 & 70.40 & $-1.21$ \\
 & MLVU           & 74.79 & 76.50 & $-1.71$ \\
 & MotionBench    & 59.56 & 61.40 & $-1.84$ \\
 & LongVideoBench & 62.90 & 66.00 & $-3.10$ \\
 & LVBench        & 45.24 & 50.90 & $-5.66$ \\
\cmidrule(lr){1-5}
\multicolumn{4}{l}{\textit{mean absolute difference}} & \textit{2.70} \\
\bottomrule
\end{tabular}
\caption{Our reproductions against the scores each release reports, on the
benchmarks that release covers.}
\label{tab:repro}
\end{table}

\end{document}